\documentclass[runningheads]{llncs}

\usepackage{eccv}

\usepackage{eccvabbrv}
\usepackage{wrapfig}
\usepackage{multirow}
\usepackage{graphicx}
\usepackage{booktabs}
\usepackage{capt-of}

\usepackage[accsupp]{axessibility}  

\usepackage{hyperref}

\usepackage{orcidlink}

\definecolor{ben}{rgb}{0.9,0.,0.5}

\newcommand{\OURS}{\mbox{ReCaLaB}}

\begin{document}

\title{Reality's Canvas, Language's Brush: Crafting 3D Avatars from Monocular Video}

\titlerunning{ReCaLab}

\author{{Yuchen Rao\inst{1} \and
Eduardo Perez Pellitero\inst{1} \and
Benjamin Busam\inst{2}} \and \\
{Yiren Zhou\inst{1} \and Jifei Song\inst{1}}}


\authorrunning{Y. Rao et al.}

\institute{\small$^1$Huawei London Research Center\quad$^2$Technical University of Munich \\
\email{yuchen.rao@huawei.com\quad  e.perez.pellitero@huawei.com\quad}  \\
\email{b.busam@tum.de\quad
 zhouyiren@huawei.com\quad  jifeisong@huawei.com}}

\maketitle
\begin{abstract}

Recent advancements in 3D avatar generation excel with multi-view supervision for photorealistic models.
However, monocular counterparts lag in quality despite broader applicability.
We propose \OURS{} to close this gap.
\OURS{} is a fully-differentiable pipeline that learns high-fidelity 3D human avatars from just a single RGB video.
A pose-conditioned deformable NeRF is optimized to volumetrically represent a human subject in canonical T-pose.
The canonical representation is then leveraged to efficiently associate neural textures using 2D-3D correspondences.
This enables the separation of diffused color generation and lighting correction branches that jointly compose an RGB prediction.
The design allows to control intermediate results for human pose, body shape, texture, and lighting with text prompts.
An image-conditioned diffusion model thereby helps to animate appearance and pose of the 3D avatar to create video sequences with previously unseen human motion.
Extensive experiments show that \OURS{} outperforms previous monocular approaches in terms of image quality for image synthesis tasks. Moreover, natural language offers an intuitive user interface for creative manipulation of 3D human avatars.
\end{abstract}    
\section{Introduction}
\label{sec:intro}

\begin{figure}
\centering
\includegraphics[width=\textwidth]{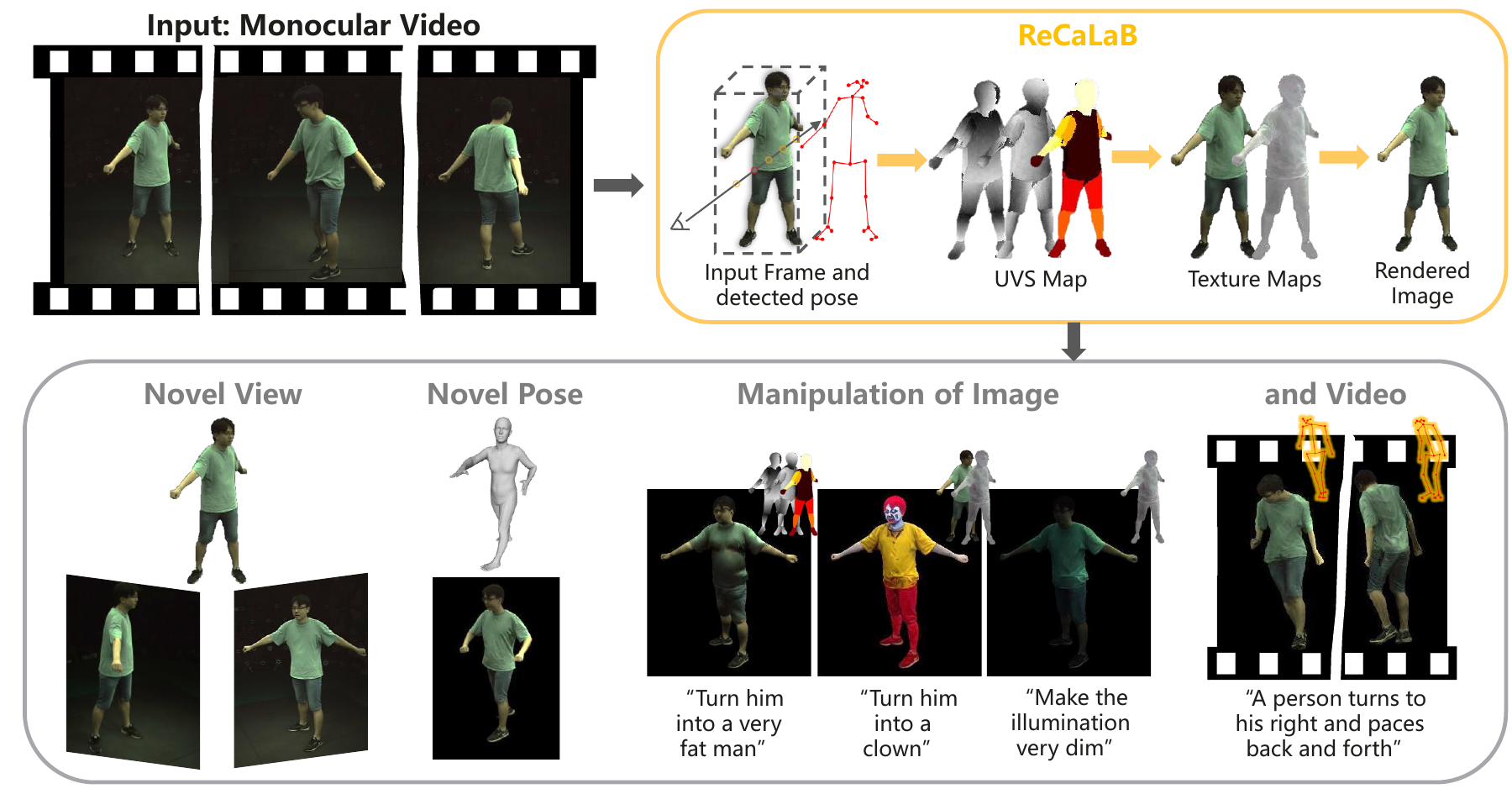}
\caption{\textbf{Overview of \OURS{}.} We propose \OURS{}, a novel approach for photo-realistic 3D avatar creation from monocular video and different types of manipulations. Our approach models neural humans with a disentangled neural texture field and enables different types of manipulation, including novel view rendering, novel pose animation, and novel appearance with customized shape, texture, and illumination by language brush.} 
\label{fig:teaser}
\vspace{-5mm}
\end{figure}

In an age of digital immersion, the demand for realistic and adaptable 3D avatars is more pronounced than ever. With the prevalent use of webcams and video calls, personalizing digital human interaction in 2D has become commonplace. The natural progression now calls for avatars to transcend into the realm of 3D, representing us in the digital space. Approximate 3D human appearance~\cite{loper2015smpl,pavlakos2019smplx,osman2020star} and body part models~\cite{li2017learning,romero2022embodied} can already be estimated in monocular RGB~\cite{kocabas2019vibe,sun2021monocular}, video~\cite{sun2023trace} and multi-view settings~\cite{dong2021shape,bastian2023disguisor}.

Traditionally, creating high-fidelity 3D avatars was a resource-intensive process, often requiring skilled artists to invest hours animating a single individual. Recent advancements~\cite{mildenhall2021nerf,pumarola2021d,li2022neural,chen2023uv}, while revolutionizing photorealistic representations, have relied on costly multi-view synchronized video capture~\cite{ionescu2013human3p6,peng2021neural,icsik2023humanrf}, constraining their accessibility and practicality for everyday use. This limitation raises a pivotal question: how can we democratize the creation of intricate 3D avatars, making it accessible to a wider audience through a seamless manipulation interface?

Enter ReCaLaB, a gateway where monocular 3D avatar creation converges with an intuitive language interface. Leveraging a single monocular video, we produce a digital 3D human that can be effortlessly manipulated with a user text prompt. This liberates avatar creation from the confines of specialized lab setups, eliminating the need for hardware expertise to construct complex camera arrays while expert coding knowledge is no longer required to manipulate learned digital representations. The title ``Reality's Canvas, Language's Brush'' encapsulates this vision, where language acts as the brushstroke that brings avatars to life on the canvas of real monocular video data, fostering creative expression on a global scale and empowering a broader community.

Technically, we leverage the recent progress in implicit 3D neural radiance fields~\cite{mildenhall2021nerf} and their deformable counterparts targeting 3D human generation~\cite{jiang2022neuman,weng2022humannerf,chen2023uv} to generate a 3D avatar. Our pipeline \OURS{} is illustrated in Fig.~\ref{fig:overview}.
At its core, \OURS{} employs a fully-differentiable pipeline. Leveraging the seminal ideas of HumanNeRF~\cite{weng2022humannerf}, we learn a deformation field from a monocular video that transports pixel input into a canonical 3D representation conditioned on the human skeletal motion in the video frames.
Using learned 3D-2D correspondences from this canonical T-pose, we generate viewpoint-agnostic neural texture stacks~\cite{chen2023uv} to represent the diffused color, along with the lighting correction to efficiently compose a final 2D rendering. 
A textual prompt can then be leveraged by a diffusion model~\cite{tevet2023human,brooks2023instructpix2pix} to generate either a human motion or change the geometry and appearance of the 3D avatar. The fully-differentiable nature of this pipeline thereby allows to propagate gradients of an image-conditioned diffusion model~\cite{brooks2023instructpix2pix} for a targeted change of individual components in \OURS{}. Controlled un-freezing of intermediate parts enables changing body shape, texture, or lighting while being free to choose the camera viewpoint.

The result is a model that not only outperforms existing monocular approaches in terms of image quality but also surpasses multi-view methods, even when the latter leverages a significantly greater volume of data.

In summary, our contributions are threefold:
\begin{enumerate}
\item \textbf{Differential Avatar Manipulation}: A fully-differentiable pipeline enables targeted \textbf{control} of appearance and geometry \textbf{through text input}. Intermediate representations encompass key aspects such as body shape, pose, texture, lighting, and camera viewpoint.
\item A \textbf{Viewpoint-agnostic Canonical Representation} of 2D diffused textures and 2D-3D correspondences allows to learn high-fidelity 3D textures from a \textbf{single RGB video} source. Resulting in rendering quality under novel viewpoints and human poses outperforms even multi-view approaches. 
\item A \textbf{Decomposed Appearance Generation} module discerns diffused color and lighting correction branches based on the pose and view-direction predictions. This enables modification of illumination while preserving texture integrity.
\end{enumerate}

\section{Related Works}
\subsection{3D Human Avatars}
In the realm of 3D human representation, progress has been notable, with early methods focusing on predicting keypoints and skeletons~\cite{openpose,insafutdinov2016deepercut}, later evolving to incorporate volumetric details from data~\cite{loper2015smpl} including details for hands, face, and feet~\cite{SMPL-X:2019,osman2020star}. Dataset acquisition has seen substantial efforts since early multi-view reconstruction works~\cite{kanade1997virtualized}, with advancements in hardware setups and accurate annotations~\cite{ionescu2013human3p6,peng2021neural,zheng2022structured,icsik2023humanrf}. However, obtaining high-quality 3D annotations remains a costly endeavor~\cite{wang2022phocal,jung2022housecat6d,icsik2023humanrf}.
In the pursuit of animatable human avatars, the field has progressed from static 3D surface recovery~\cite{saito2019pifu,saito2020pifuhd} to sophisticated neural approaches guided by 3D human models.
The increasing popularity of implicit scene representations~\cite{Mescheder2019occupancy,Park2019deepsdf}, in particular static neural fields~\cite{mildenhall2021nerf,muller2022instant,barron2022mip} and their deformable counterparts~\cite{li2022neural,pumarola2021d} have accelerated this process even from monocular video input~\cite{park2021nerfies,park2021hypernerf,li2021neural}.
The emergence of neural rendering methods, including Neural Body~\cite{pumarola2021d}, Neural Articulated Radiance Field~\cite{noguchi2021neural}, A-NeRF~\cite{su2021nerf}, TAVA~\cite{li2022tava}, and PoseVocab~\cite{li2023posevocab}, signify a significant shift towards learning human shape, appearance, and pose from multi-view data. These methods allow to synthesize photorealistic images from novel 3D viewpoints and to animate human characters with given poses.
Additionally, for monocular video, approaches like SelfRecon~\cite{jiang2022selfrecon} and NeuMan~\cite{jiang2022neuman} as well as Animatable NeRF~\cite{peng2021animatable} and HumanNeRF~\cite{weng2022humannerf} integrate motion priors for regularization to allow this even with a single input video.
While these innovations excel in rendering quality, they come with the challenge of high computational demands during both training and testing phases. Concurrent to our approach, explicit methods using deformable 3D Gaussian splatting~\cite{kerbl20233d} is also explored for 3D avatar generation~\cite{jung2023deformable,moreau2024human,kocabas2023hugs}. Both directions, however, lack the capability for intuitive texture editing due to their 3D scene representation. Our pipeline also generates a detailed human avatar from just a single monocular video, but additionally offers an intuitive way to manipulate its texture, pose, and shape through text prompts. This unique capability is made possible by leveraging 2D-3D correspondences.

NeRFs excel in reproducing visual appearance even with fixed geometry~\cite{chen2023texpose}. While previous approaches improve texture fidelity through adversarial training~\cite{wang2021towards} or latent diffusion~\cite{svitov2023dinar}, correspondences maps to 3D human models in the form of SMPL~\cite{loper2015smpl} or DensePose~\cite{guler2018densepose} are a nifty alternative.
UV Volumes~\cite{chen2023uv} leverages this strength to create an efficient rendering pipeline for human avatars, utilizing 2D-3D correspondences, a technique also employed in other domains~\cite{li2023nerf}. This significantly accelerates both learning and inference processes while allowing to directly manipulate texture in 2D. The method shows impressive results when being trained with multi-view data.
We also leverage a correspondence map in the form of 24 human body segments, and learn a canonical UV-map together with a neural texture stack to capitalize on information from just a monocular video mapped into a canonical space.

\subsection{Manipulation with Diffusion Models}
Diffusion models, a class of generative models employing stochastic diffusion~\cite{sohl2015deep,song2020improved}, have shown promise in various applications. During training, noise is gradually introduced to a sample from the data distribution, while the inverse process is learned starting from pure initial noise. This technique has been extended for image generation~\cite{ho2020denoising,song2020denoising}, mesh editing~\cite{michel2022text2mesh, richardson2023texture}, and in recent years, conditioned diffusion models have gained traction in text-guided image generation~\cite{dhariwal2021diffusion,nichol2021glide,ho2022classifier}, where an image guides the denoising process.
In the realm of human motion, diffusion models have been leveraged to generate complex motion sequences for human avatars~\cite{Yu_2023_CVPR}. We adopt a similar diffusion model~\cite{tevet2023human} to generate text-driven motion sequences within \OURS{}.

Noteworthy examples in human image generation include DreamPose~\cite{karras2023dreampose}, which creates synthetic fashion videos guided by still images and pose sequences, and ControlNet~\cite{zhang2023adding}, capable of generating human images based on novel poses and textual descriptions. Works like Dreamface~\cite{zhang2023dreamface}, TeCH~\cite{huang2023tech}, and~\cite{kolotouros2024dreamhuman, aneja2023clipface, hong2022avatarclip, wang2023rodin, wu2023high} employ diffusion models to learn editable textures for synthetic human faces and bodies.
Moreover, InstructPix2Pix~\cite{brooks2023instructpix2pix} stands out for its impressive results, and its extension, Instruct-NeRF2NeRF~\cite{haque2023instruct}, ensures frame consistency in static real-world scenes. In our approach, we apply a similar diffusion process to a deformable scene mapped to a canonical parametric human representation, allowing us to edit the UVS map or neural texture stack conditioned on a text prompt.

\begin{figure*}[!hbt]
\centering
\includegraphics[width=\textwidth]{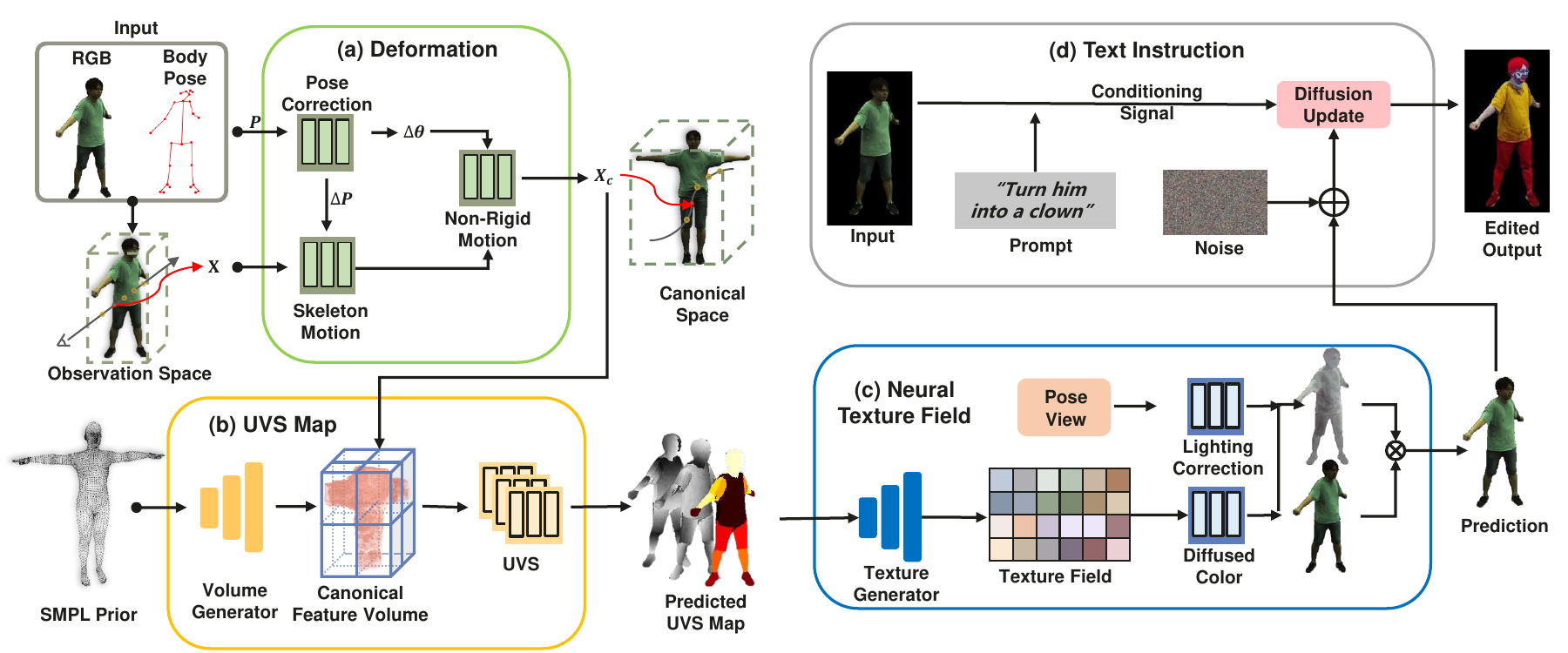}
\vspace{-3mm}
\caption{\textbf{Framework of \OURS{}.} Our approach takes monocular video frames as the input to reconstruct the human avatar. We first learn the backward deformation $T$ of the human body in module (\textit{a}) with the human pose. Then in module (\textit{b}), we generate a corresponding UVS map by using a volume generator $G_c^F$ initialized by the SMPL T-pose prior, followed with different mapping functions $G_c^M, (M \in \{U, V, S\})$. We then designed a texture generator $H_c^T$ along with the diffused color branch $H_c^A$, and the lighting correction branch $H_c^N$ for the neural texture learning in module (\textit{c}). Finally, in module (\textit{d}), shape, texture, and lighting correction modules can be updated accordingly with the instruction from the language brush.}
\label{fig:overview}
\end{figure*}

\section{Method}

Given a monocular video that captures the human body, we propose an efficient technique to reconstruct the 3D human avatar, which can be further driven by a single-sentence instruction. The overview of our method is shown in Fig.~\ref{fig:overview}, where our approach consists mainly of three fundamental components: 1) we learn a hybrid neural representation to model the human avatar, with a neural feature volume field and neural texture extraction. 2) We introduce a lighting correction network to refine the learned neural texture to cope with lighting differences. 3) In the end, we update the editable neural human representation with language instruction as a brush.
With our full pipeline, we can lift the human into a 3D editable neural avatar from a monocular human video alone and edit it with human language descriptions. 

\subsection{Reality's Canvas: Dynamic Human Modeling}

We first define a canonical neural field to represent the human avatar in a canonical space:

\begin{equation}
F_c:(\mathbf{x}_c, \mathbf{P}, \mathbf{d}) \mapsto \mathbf{c},
\end{equation}
where $\mathbf{x}_c$ is a sampled 3D point from the human body in canonical space, $\mathbf{P}$ denotes the human pose for the given time frame, and $\mathbf{d}$ represents the viewing direction. $\mathbf{c}$ is the mapped color in texture space output by the mapping function, $F_c$. 

The mapping function $F_c$ can be further decomposed into shape reconstruction with UV coordinates, 3D semantic estimation, and in the end, texture synthesis with UVS mapping. In this decomposition, We first learn a shared feature representation generated by the neural feature volume generator, $G_c^F$:

\begin{equation}
G_c^F:\mathbf{x}_c \mapsto \mathbf{f}
\end{equation}
where $G_c^F$ serves as the mapping function to take the 3D input position $\mathbf{x}_c$ to obtain the $\mathbf{f}$ to form the neural feature volume.

Following that, We first decompose the 3D shape information from the shared latent space for modeling the human body with an implicit neural representation and represent it in UV coordinates: 

\begin{equation}
\label{eq:uv}
G_c^U:\mathbf{f} \mapsto \mathbf{u}, \; \; G_c^V:\mathbf{f}  \mapsto \mathbf{v}
\end{equation}
where $G_c^U$ and $G_c^V$ are the mapping function to take the 3D feature $\mathbf{f}$ to obtain the $\mathbf{u}$ and $\mathbf{v}$ information in the neural volume, representing the human shape. Similarly, we also formalize the human semantics in the neural volume following \cite{Zhi:etal:ICCV2021} with mapping function, $G_c^S$:

\begin{equation}
\label{eq:s}
G_c^S:\mathbf{f} \mapsto \mathbf{s}
\end{equation}
To model the dynamics in human motion and animate the lifted 3D human avatar, we further introduce the deformation module, $T$, to connect the 3D points $\mathbf{x}$ in the observation space, with the one in the canonical space, $\mathbf{x}_c$. In other words, the mapping function in the observation space is connected to the aforementioned mapping function in canonical space via $T$. Thus, we can obtain the neural volume in observation space, $G^S$, by warping the canonical neural volume:

\begin{equation}
\left(G_c^S \circ G_c^F\right) \big(T(\mathbf{x}, \mathbf{P}, \mathbf{d})\big),
\end{equation}
We apply the same deformation field to bend the rays of the mapping function for UV representations. Volume rendering is then applied to the output $\mathbf{U}$, $\mathbf{V}$ and the semantic map $\mathbf{S}$. 

Inspired by HumanNeRF~\cite{weng2022humannerf}, our final deformation, $T$, is defined as the combination of rigid skeleton motion depending on the corrected body pose and a residual non-rigid motion. With the human motion-based deformation module, the network will be able to understand the human body motions correctly and make a transition from a given 3D point in observation space into the canonical space. 

\subsection{Disentangled Neural Texture Learning}

With the rendered UV map and semantic map, our network can further synthesize high-quality neural textures. We realize this by decomposing appearance generation into diffused color and lighting correction components within a disentangled neural texture learning module. Specifically, we build another neural mapping function, $H_c^T$, to generate the neural texture field in the texture space with regard to the input UV and semantic space:

\begin{equation}
H_c^T:(\mathbf{U}, \mathbf{V}, \mathbf{S}) \mapsto \mathbf{t}.
\end{equation}
where $\mathbf{U}$, $\mathbf{V}$, and $\mathbf{S}$ are the UV map and semantic representation and are taken as input to the mapping function, $H_c^T$, to output the neural texture representation, $\mathbf{t}$, to form the neural texture field. 

Compared to learning a neural radiance field to synthesize the color of the human avatar, learning a neural texture will produce better appearance modeling by avoiding all the floaters above the texture surfaces. 
However naive neural texture cannot reflect the lighting change affected by occlusion, viewpoint, or human pose alteration. To cope with this issue, we then decompose the neural texture representation into diffused color and lighting correction branches. Firstly, we use MLPs to convert the neural texture representation to the RGB color component, reflecting the diffused color of the surface to represent the human appearance in the bright and evenly distributed lighting condition: 

\begin{equation}
H_c^A:\mathbf{t} \mapsto \mathbf{a}.
\end{equation}
where $\mathbf{a}$, denotes the diffused color component and, $H_c^A$, represents the mapping function in the diffused color branch.

We further encode lighting correction information to represent the illumination condition influenced by the pose articulation and view direction change:  

\begin{equation}
H_c^N:(\mathbf{t}, \mathbf{d}, \mathbf{P}) \mapsto \mathbf{m}
\end{equation}
where $H_c^N$ is the learned neural mapping function, that maps the neural texture, $\mathbf{t}$, with the guidance of view direction $\mathbf{d}$, and human poses $\mathbf{P}$, to the lighting correction information, $\mathbf{m}$. To this end, the final color for each ray in 2D image plane is calculated as $\mathbf{c} = \mathbf{a} \cdot \mathbf{m}$.

\subsection{Languages's Brush: Editing Neural Human}
\label{sec:diffusion}

Our disentangled neural texture generation allows us to manipulate texture and shape separately. This gives us explicit control over which part of the pipeline is changed for synthesis upon user inputs. Given text instruction for neural human editing, our framework can be customized to update the corresponding module accordingly related to the instruction, enabling more accurate and efficient neural human editing. 

Following the idea of Instruct-NeRF2NeRF~\cite{haque2023instruct} for scenes, we utilize the denoising diffusion implicit models (DDIM) to edit a 2D rendered image conditioned on a text description and iteratively update the image supervision to edit the entire human avatar. Our design allows us to go beyond uncontrolled manipulation to more specific editing by connecting the input text with corresponding modules. For example, if we like to change the color of the human avatar's cloth while keeping the human's identity and other features in a virtual try-on scenario, we can simply fix the latter. Since our approach has disentangled texture and shape information, we can choose to back-propagate the gradient flow only into the neural texture part to edit the texture space, while maintaining the same human shape. 

\subsection{Training Objectives}

\OURS{} is trained with a photometric reconstruction loss between the rendered color image and the ground-truth, including both mean squared error (MSE) loss and LPIPS loss: 

\begin{equation}
\label{eq:l_rec}
\mathcal{L}_{\text{rec}}=\lambda_\text{LPIPS}\mathcal{L}_\text{LPIPS} + \lambda_\text{MSE}\mathcal{L}_\text{MSE},
\end{equation}
where $\lambda_\text{LPIPS}$ and $\lambda_\text{MSE}$ are the weights for the corresponding losses. 
The photometric loss will also back-propagate to train the mapping functions, $G_c^U$, $G_c^V$, and $G_c^S$ to predict meaningful UV and semantic maps, which are further regularised by another three losses, $\mathcal{L}_{reg}^U$, $\mathcal{L}_{reg}^V$ and $\mathcal{L}_{reg}^S$ between the predictions and pseudo labels, obtained from a pre-trained Densepose\cite{guler2018densepose} network. 
In addition to the regularization losses, we add a smoothness term $\mathcal{L}_\text{smt}$ to encourage the neighbor 3D points to embed the same semantics by leveraging human parsing priors. 

\begin{equation}
\label{eq:l_reg}
\mathcal{L}_{\text{reg}}=\mathcal{L}_\text{reg}^U + \mathcal{L}_\text{reg}^V + \mathcal{L}_\text{reg}^S + \mathcal{L}_\text{smt},
\end{equation}
where both $\mathcal{L}_\text{reg}^U$ and $\mathcal{L}_\text{reg}^V$ are formulated with MSE loss, $\mathcal{L}_\text{reg}^S$ uses cross entropy loss, and $\mathcal{L}_\text{smt}$ uses $l_2$ loss. 

To reduce flying pixels and condense the learned neural volume in line with the human shape, we utilize a mask loss to constrain geometry reconstruction with $\mathcal{L}_\text{mask}$, similar to \cite{yang2021objectnerf}. Finally, we obtain the overall training objective in Eq.~\ref{eq:total_loss} by weighting all the aforementioned losses to generate the neural human avatar:

\begin{equation}
\label{eq:total_loss}
\mathcal{L}=\lambda_\text{rec}\mathcal{L}_\text{rec} + \lambda_\text{reg}\mathcal{L}_\text{reg} + \lambda_\text{mask}\mathcal{L}_\text{mask},
\end{equation}
where $\lambda_\text{rec}$, $\lambda_\text{reg}$ and $\lambda_\text{mask}$ are the loss weight for $\mathcal{L}_\text{rec}$, $\mathcal{L}_\text{reg}$ and $\mathcal{L}_\text{mask}$, respectively. 

In order to train the text editing module, we employ the predicted image as the input, utilizing the ground truth image as the original reference. Editing is performed using InstructPix2Pix~\cite{brooks2023instructpix2pix} in response to the given text prompt. To facilitate this process, the edited image is utilized for computing $\mathcal{L}_\text{rec}$, in conjunction with the predicted image.

\begin{figure*}[!hbt]
\includegraphics[width=\textwidth]{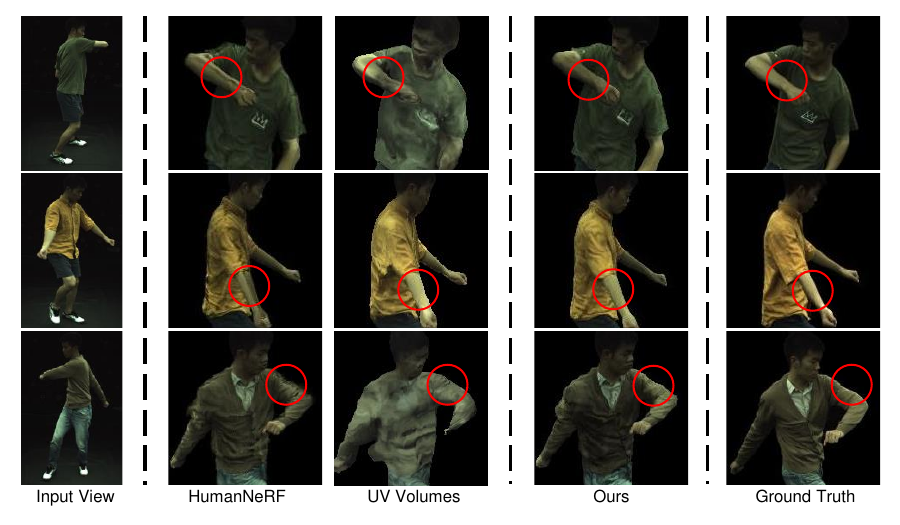}
\vspace{-5mm}
\caption{\textbf{Qualitative results for human rendering on ZJU-MoCap dataset~\cite{peng2021neural} for novel view.} HumanNeRF~\cite{weng2022humannerf} suffers from ghosting artifacts (see marked regions) while deformation artifacts arise in UV volumes~\cite{chen2023uv} due to the texture-driven training. Our approach copes with both these issues providing results visually closer to the ground truth.} 
\label{fig:zju}
\end{figure*}

\section{Experiments}


\subsection{Implementation Details}
We adopt Adam optimizer~\cite{kingma2014adam}. We set the learning rate to $5\times 10^{-4}$ in the initial stage with 200K iterations to warm up the training, where we apply the regularization loss with pseudo labels extracted by the pre-trained DensePose~\cite{guler2018densepose} network to constrain the learning of the UV map and semantic information. After the warm-up stage, we will remove the regularization loss and jointly train all modules with additional mask loss for the remaining 200K iterations with $1\times 10^{-4}$ as the learning rate. The whole training stage takes about 60 hours on a single V100 GPU. During inference, our pipeline renders 3.6s per frame with 512 $\times$ 512 image resolution. Hyper-parameters setting and other additional implementation details can be found in the supplementary.  

\subsection{Dataset and Metrics}

\paragraph{\textbf{Dataset}}
We train and evaluate our approach on the ZJU-MoCap dataset~\cite{peng2021neural} and the People Snapshot dataset~\cite{alldieck2018video}. The ZJU-Mocap dataset is captured with 23 hardware-synchronized cameras. 
For an equal comparison with HumanNeRF~\cite{weng2022humannerf}, we use the same experimental setting by splitting video frames captured by the first camera into seen and unseen pose sets in a 4:1 ratio and use other camera views with the same seen pose as the novel view evaluation set. We repeat this procedure on 6 different subjects and then average the quantitative results. For the People Snapshot dataset, we run EasyMocap~\cite{easymocap} to extract the approximate camera and body pose, and then follow the same training and testing split. 

\begin{figure}[htbp]
        \begin{minipage}[t]{0.48\textwidth}
        \centering
    \includegraphics[height=3.0cm]{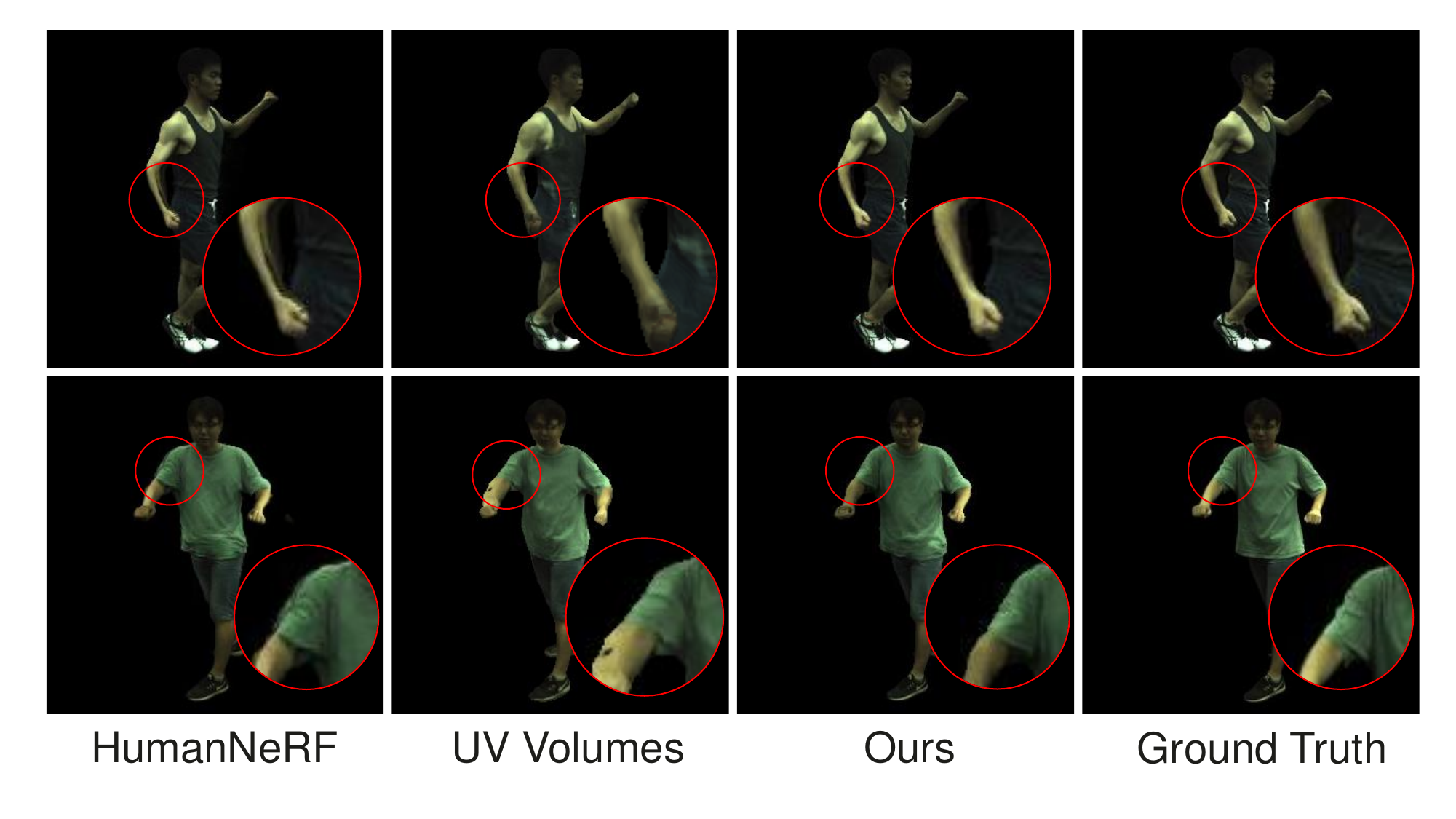}
    \caption{\textbf{Qualitative results for human rendering on ZJU-MoCap dataset~\cite{peng2021neural} for novel pose.} Our method yields the most superior texture outcomes.}
    \label{fig:novel_pose_zju}
        \end{minipage}
        \begin{minipage}[t]{0.02\textwidth}
        ~~
        \end{minipage}
        \begin{minipage}[t]{0.48\textwidth}
        \centering
    \includegraphics[height=3.2cm]{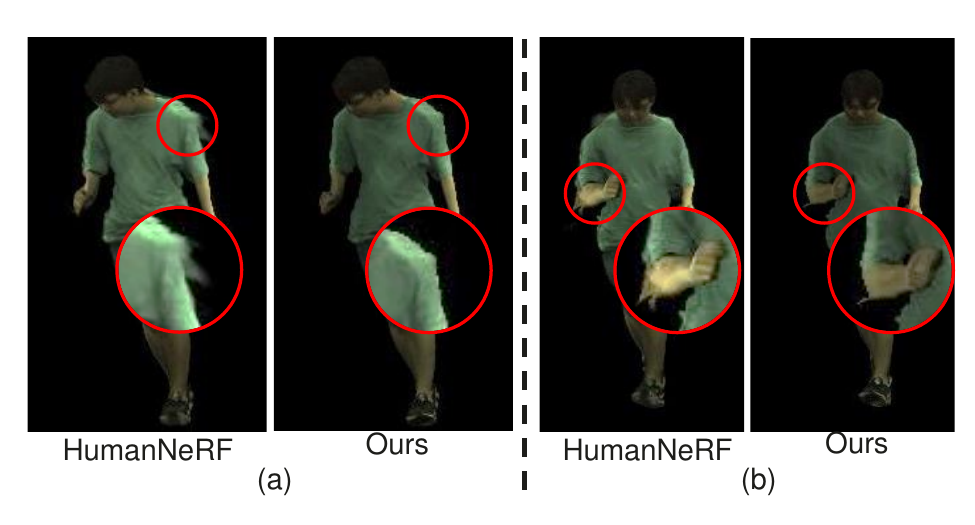}
    \caption{\textbf{Qualitative results on novel poses guided by text instruction.} \OURS{} provides sharper texture and fewer geometric artifacts compared to HumanNeRF~\cite{weng2022humannerf}.}
    \label{fig:novel_pose}
        \end{minipage}
\end{figure}

\paragraph{\textbf{Evaluation Metrics}}
In line with preceding neural human approaches, we assess the quality of rendering neural human with the following evaluation metrics: the peak signal-to-noise ratio (PSNR), the structural similarity index (SSIM), and the learned perceptual image patch similarity (LPIPS).

\subsection{Comparison with State-of-the-art Methods}

\paragraph{\textbf{Baselines}}

We compared our method with state-of-the-art methods for neural human rendering, including both HumanNeRF~\cite{weng2022humannerf} and UV Volumes~\cite{chen2023uv}. We obtain all the baseline results using the official implementations under the same experimental setting. 

\paragraph{\textbf{Comparison on Novel View Synthesis}}
We first make comparisons of the rendering performance for novel view synthesis, where we synthesize all the hold-out views under the seen poses. Tab.~\ref{tab:zju_all} summarized our performance compared to state-of-the-art (SotA) baselines. Compared to HumanNeRF~\cite{weng2022humannerf}, our method achieves competitive performance on PSNR, while achieving better performance on SSIM and LPIPS metrics. Moreover, our method outperforms UV Volumes~\cite{chen2023uv} on all the evaluation metrics, showing that our method is effective in modeling the neural human from even a single camera. Qualitatively, as highlighted in the visualization from Fig.~\ref{fig:zju}, our method also synthesizes more detailed texture thanks to the proposed disentangled neural texture learning. More examples can be found in the supplementary.

\begin{table}[htbp]
        \begin{minipage}[t]{0.48\textwidth}
        \centering
    \caption{Quantitative results for  comparing to baselines on ZJU-Mocap dataset~\cite{peng2021neural}.}
\label{tab:zju_all}
\resizebox{1.0\columnwidth}{!}{
\begin{tabular}{lcccccc}
\toprule 
\multirow{1}{*}{} & \multicolumn{3}{c}{\textbf{Novel View}} & \multicolumn{3}{c}{\textbf{Novel Pose}} \tabularnewline
\cmidrule(lr){2-4}\cmidrule(lr){5-7}
 & PSNR $\uparrow$ & SSIM $\uparrow$ & LPIPS $\downarrow$ & PSNR $\uparrow$ & SSIM $\uparrow$ & LPIPS $\downarrow $  \tabularnewline
\midrule 
UV Volumes\cite{chen2023uv} & 23.31 & 0.962 & 0.047 & 23.06 & 0.968 & 0.040 \\
HumanNeRF\cite{weng2022humannerf}  & \textbf{29.57} & 0.967 & 0.030 & 29.62 & 0.969 & 0.029 \\
Ours & 29.40 & \textbf{0.968}& \textbf{0.029} & \textbf{29.66} & \textbf{0.970} & \textbf{0.028} \\
\hline
\end{tabular}
}
        \end{minipage}
        \begin{minipage}[t]{0.02\textwidth}
        ~~
        \end{minipage}
        \begin{minipage}[t]{0.48\textwidth}
        \centering
\caption{Quantitative results for comparing to UV Volumes~\cite{chen2023uv} with different number of training views.}
\label{tab:diffview}
\resizebox{1.0\columnwidth}{!}{
\begin{tabular}{lcccccc}
\toprule 
\multirow{1}{*}{} & \multicolumn{3}{c}{\textbf{Novel View}} & \multicolumn{3}{c}{\textbf{Novel Pose}} \tabularnewline
\cmidrule(lr){2-4}\cmidrule(lr){5-7}
 & PSNR $\uparrow$ & SSIM $\uparrow$ & LPIPS $\downarrow$ & PSNR $\uparrow$ & SSIM $\uparrow$ & LPIPS $\downarrow $  \tabularnewline
\midrule 
UVV-mono\cite{chen2023uv} & 24.13 & 0.967 & 0.040 & 23.47 & 0.969 & 0.042 \\
UVV-full\cite{chen2023uv}  & 27.63 & \textbf{0.981} & \textbf{0.024} & 23.80 & \textbf{0.972} & 0.037 \\
Ours & \textbf{29.25} & 0.969 & 0.027 & \textbf{30.36} & \textbf{0.972} & \textbf{0.026} \\
\hline
\end{tabular}
}
        \end{minipage}
\end{table}

\paragraph{\textbf{Comparison on Novel Pose Rendering}}
We compare our method against the baselines on novel pose rendering. As listed in Tab.~\ref{tab:zju_all}, our method shows the best performance for the novel pose scenario with a large margin over UV Volumes~\cite{chen2023uv}, and Fig.~\ref{fig:novel_pose_zju} indicates that our method provides better texture and fewer geometric artifacts compared to other baselines.

\begin{figure*}[!hbt]
\centering
\vspace{-5mm}
\includegraphics[width=\textwidth]{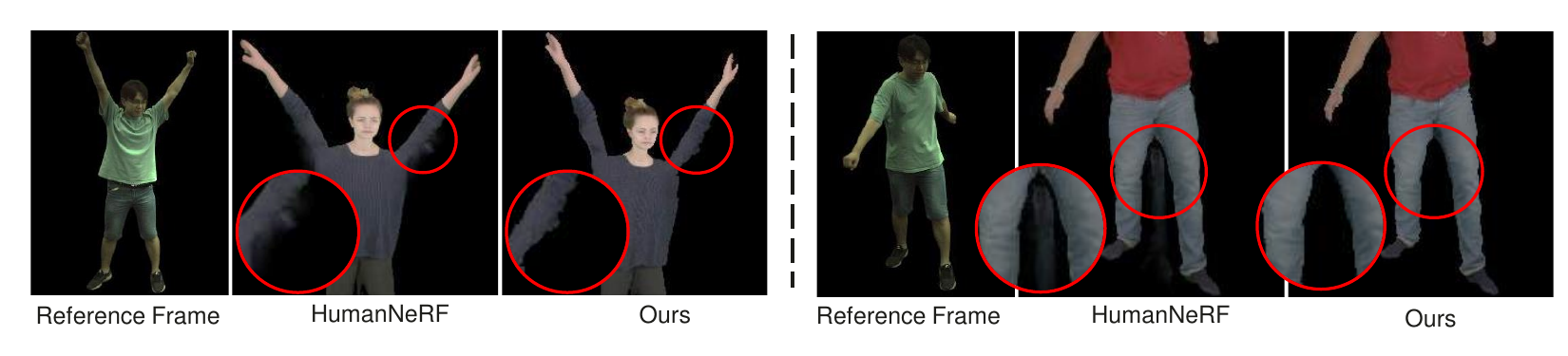}
\caption{\textbf{Qualitative results on People Snapshot dataset~\cite{alldieck2018video}.} Reconstructed Human Avatar for both \textit{female-3-casual} and \textit{male-3-casual} are rendered under novel pose which is triggered by the reference frames. \OURS{} provides less artifacts and visually more pleasing results compared to HumanNeRF~\cite{weng2022humannerf}.}
\label{fig:peoplesn}
\vspace{-5mm}
\end{figure*}

We further present a qualitative comparison between our method and HumanNeRF~\cite{weng2022humannerf} in Fig.~\ref{fig:novel_pose}, where the neural avatar is driven by novel poses generated by text instruction with the help of MDM~\cite{tevet2023human}, a) and b) are frames from the instructions \textit{``A person turns to his right and paces back and forth''}, and \textit{``A person is running''} respectively. We observe that our method can render the human avatar with better quality and fewer artifacts. Similarly, qualitative results on the People Snapshot dataset~\cite{alldieck2018video} in Fig.~\ref{fig:peoplesn} illustrate that \OURS{} avoids floating artifacts and animates the neural human avatar in high quality. Multi-view methods such as UV Volumes~\cite{chen2023uv} heavily depend on input from multiple synchronized cameras which hinders their ability to achieve compelling rendering results in single-view scenarios for novel poses beyond the learned pose domain. We observed that UV Volumes~\cite{chen2023uv} fails to produce adequate results in this challenging setup.

\begin{figure*}[!hbt]
\centering
\includegraphics[width=\textwidth]{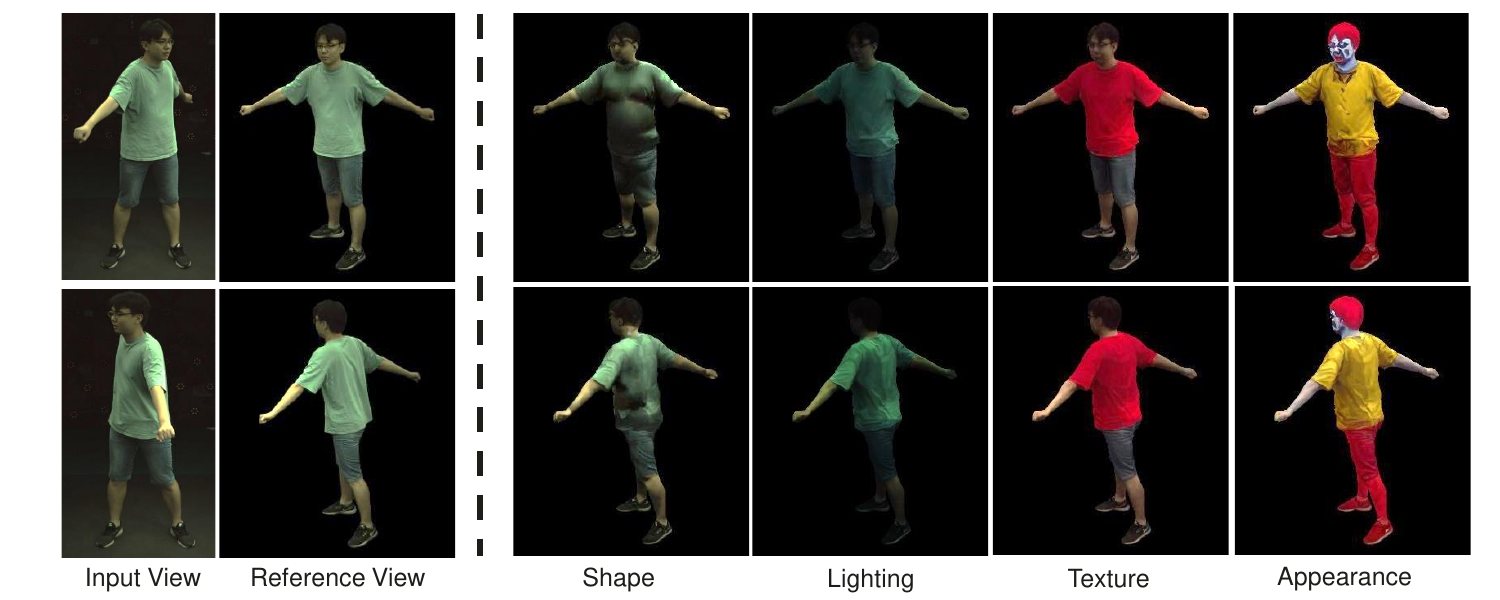}
\vspace{-4mm}
\caption{\textbf{Qualitative results for text instructed human rendering on ZJU-MoCap dataset~\cite{peng2021neural} for novel views.} Our approach is capable of generating a consistent human body based on different text instructions. More specifically, our approach enables fine-grained manipulation with the specific module, \ie, shape editing with UVS module, re-lighting with lighting correction branch, $H_c^N$, and in the end, texture and appearance editing with our neural texture module.}
\label{fig:in2n}
\vspace{-5mm}
\end{figure*}

\paragraph{\textbf{Qualitative Results for Texture Editing}}
\OURS{} enables editing neural humans with text instructions. For example, with text instructions like ``Turn him into a very fat man'', ``Make the illumination very dim'', ``Turn his T-shirt red'', and ``Turn him into a clown'', our approach connects the text prompt with the corresponding module and updates the editable neural texture to generate the manipulated neural avatar, as shown in Fig.~\ref{fig:in2n}.

\begin{figure*}[!hbt]
\centering
\includegraphics[width=\textwidth]{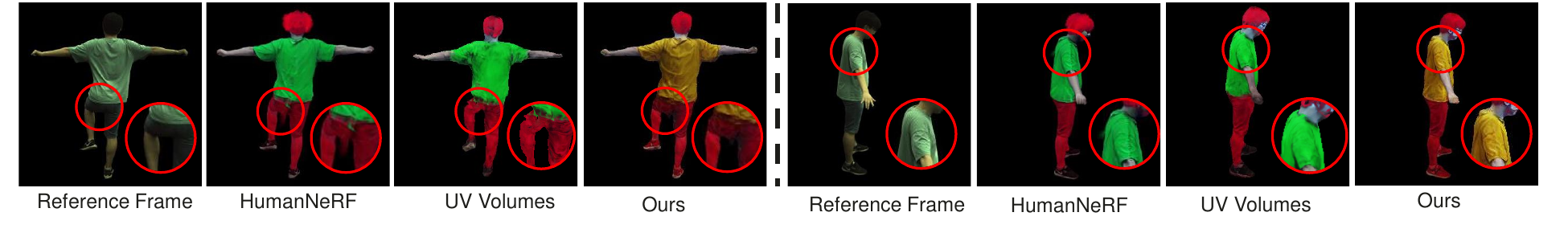}
\vspace{-5mm}
\caption{\textbf{Qualitative results on texture editing.} Our method excels in producing the best edited human avatars compared to alternative approaches.}
\label{fig:suppl_in2n}
\vspace{-5mm}
\end{figure*}

\paragraph{\textbf{Comparisons for Texture Editing}} 
We form human avatar manipulation baselines by combining HumanNeRF~\cite{weng2022humannerf} and UV Volumes~\cite{chen2023uv} with
InstructNeRF2NeRF~\cite{haque2023instruct}, and then update the canonical neural rendering module according to the text instruction. Fig.~\ref{fig:suppl_in2n} shows the comparison of the editing performance under the novel pose scenario. Specifically, the reconstructed human avatars from all the methods are edited by the text ``turn him into a clown'', and rendered under a novel pose which is triggered by the reference frame. As we can see from Fig.~\ref{fig:suppl_in2n}, our method provides visually higher quality results with fewer artifacts compared to other baselines, as highlighted with the red circles. It is also important to note that a HumanNeRF-based manipulation baseline suffers both from a considerably longer 16x training time and lower performance, due to a less-efficient patch-based training scheme.

\subsection{Ablation Studies}

\begin{figure*}[!hbt]
\centering
\vspace{-5mm}
\includegraphics[width=\textwidth]{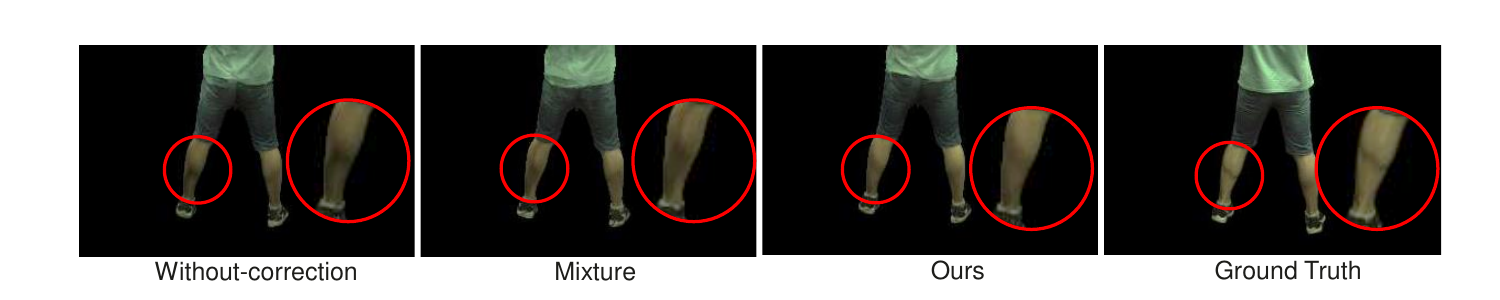}
\vspace{-5mm}
\caption{\textbf{Qualitative results for ablation on Neural Texture Field.} Our approach has achieved the best recognition of the light situation at the calf.}
\label{fig:abl}
\vspace{-7mm}
\end{figure*}

\paragraph{\textbf{Ablation on Neural Texture Field}}
We first give an ablation study on the neural texture field, where we investigate the effect of our module on modeling illumination for more photo-realistic rendering. We explore several alternative ways to our approach where we disentangle diffused color and lighting correction branches, including learning neural texture with 1) diffused color branch only (Without-correction), and 2) without disentangling the diffused color and lighting correction branches (Mixture). Upon comparison with the alternative methods, our approach stands out in effectively representing the lighting conditions, as highlighted within the red circle in Fig.~\ref{fig:abl}. 
This improvement can be attributed to our ability to proficiently learn the canonical texture while simultaneously modeling the diffused color and the lighting correction information to simulate different lighting effects based on varied view directions and human poses.

We further include the detailed quantitative comparison in Tab.~\ref{tab:ablation-texture} under novel pose synthesis for three random subjects, where our full model can achieve the best PSNR over all the alternatives, demonstrating that our approach can model the texture as well as the illumination effectively. Specifically, both our full model and the ablated version with the diffused color branch can generate sharper texture over the baseline model without disentanglement, reflected by the LPIPS metric.

\paragraph{\textbf{Ablating Current Multi-View Baseline}}
To showcase the generalization ability of \OURS{} for novel human pose rendering, we additionally include a comparison with the UVS-map driven multi-view model UV Volumes~\cite{chen2023uv} where all views are used for training. UVV-full denotes a training of~\cite{chen2023uv} with all 20 views, while the same model is trained with a single view in UVV-mono. Tab.~\ref{tab:diffview} provides the quantitative evaluation between our method \OURS{} and these two baselines by evaluating on the same held-out view (\textit{View 5}). The results demonstrate that our method is on par with UVV-full for novel view synthesis performance. For novel pose rendering, our approach even surpasses the multi-view approach and improves PSNR by approximately 6dB, even though being trained with 19x less data and no synchronized videos. These results show the effectiveness of our method by leveraging human pose priors to map geometry and texture information into a canonical neural texture field.

\paragraph{\textbf{Prompt Impact on Neural Texture Components}}
For the avatar manipulation task, we study how the neural texture field can influence the editing performance for a given text instruction, \eg, ``Make the illumination very dim''. We free the parameters of different parts of our neural texture field to make it editable while keeping the other parts frozen. The text-prompted manipulations are shown in Fig.~\ref{fig:shading}. It becomes evident that the lighting correction branch has the most significant influence on the requested illumination change by the text prompt.
As we can see, making the lighting correction branch editable satisfies most with regard to the illumination editing requested in the language description, explaining that the lighting correction branch encodes the most illumination-related information and is capable of editing on illumination for neural humans. 

\begin{table}[ht]
\vspace{-5mm}
\begin{minipage}[t]{0.48\textwidth}
\centering
\caption{Quantitative results for comparing to the ablation on neural texture field under novel pose scenario.}
\vspace{-2mm}
\label{tab:ablation-texture}
\resizebox{1.0\columnwidth}{!}{
\begin{tabular}{lccc}
\toprule 
\multirow{1}{*}{} & \multicolumn{3}{c}{\textbf{Novel Pose}} \tabularnewline
\cmidrule(lr){2-4}
 & PSNR $\uparrow$ & SSIM $\uparrow$ & LPIPS $\downarrow$ \tabularnewline
\midrule 
Without-correction & 29.77 & \textbf{0.972} & \textbf{0.024} \\
Mixture & 30.03 & \textbf{0.972} & 0.029 \\
Ours & \textbf{30.05} & \textbf{0.972} & \textbf{0.024}\\
\hline
\end{tabular}
}
\end{minipage}\hfill
\begin{minipage}[t]{0.48\textwidth}
\centering
\strut\vspace*{-\baselineskip}\newline\includegraphics[width=\textwidth]{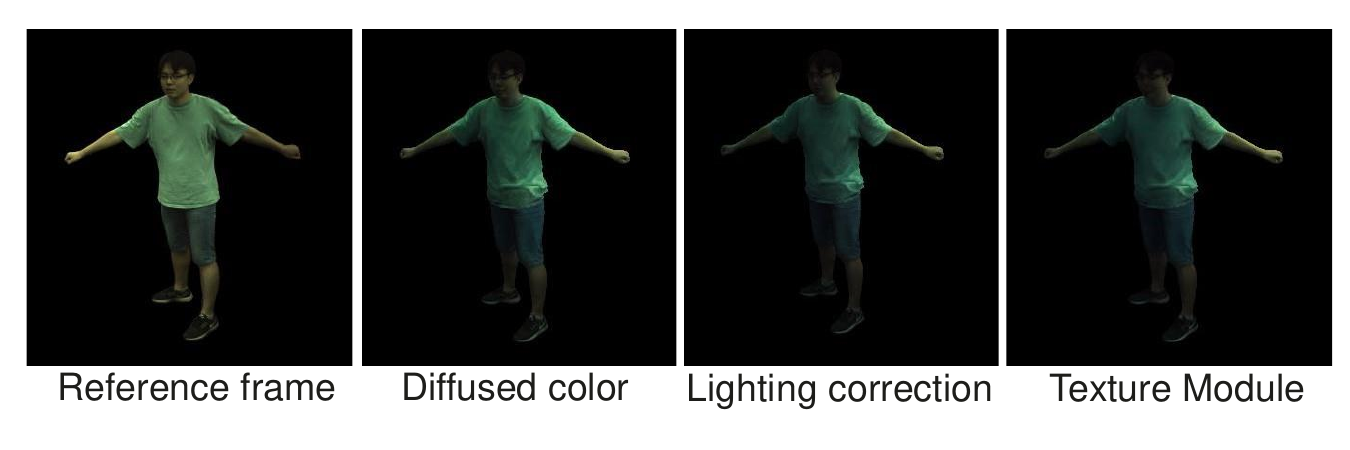}
  
  \captionof{figure}{\textbf{Qualitative results for illumination change text prompt.}  effect of the text instruction ``Make the illumination very dim'' is shown by updating only selected parts of \OURS.}
  \label{fig:shading}
\end{minipage}
\vspace{-10mm}
\end{table}

\subsection{Limitations}
Our method has shown state-of-the-art performance on both novel view synthesis and novel pose rendering making it a video-driven avatar generation pipeline on reality's canvas that can truly be edited with language’s brush. However, our method still has a few limitations: 1) Our deformation module relies on the human pose prior, thus may suffer from incorrect initial pose estimates, even though the residual pose correction module can relieve the issue to some extent. 2) Facial rendering and hand animation are significantly important for photorealistic human avatars. Despite being applicable to texture learning, \OURS{} is not yet extended to these fields. We believe that our idea can pave the way to more controllable manipulation setups there, too.

\section{Conclusion}

In this paper, we propose \OURS{}, a novel approach designed to acquire an editable neural texture field for human body rendering. Our method offers an efficient methodology of disentangled neural texture learning which can be further edited through textual instructions, making it particularly suitable for monocular video applications.
\OURS{} excels in learning a disentangled neural texture field, yielding exceptional performance in human rendering for both novel view and novel pose scenarios overcoming even recent multi-view setups with just a monocular video input that can be easily captured by everyone. Our differential avatar manipulation pipeline further democratizes human avatar manipulation through a simple language-based interface. We firmly believe that \OURS{} represents a substantial step forward in the domain of 3D avatar creation and manipulation.


%
%
\bibliographystyle{splncs04}
\bibliography{main}

\clearpage
\appendix
\setcounter{page}{1}

\setcounter{section}{0}
\setcounter{equation}{0}
\setcounter{figure}{0}
\setcounter{table}{0}
\renewcommand*{\thesection}{S\arabic{section}}
\renewcommand{\theequation}{S\arabic{equation}}
\renewcommand{\thefigure}{S\arabic{figure}}
\renewcommand{\thetable}{S\arabic{table}}

We provide additional materials to supplement our main paper. In Sec.~\ref{sec:aer}, we provide more experiments to give more detailed comparisons between our approach and the baseline methods. Furthermore, in Sec.~\ref{sec:aid}, we include addition implementation details for our approach. In the end, we also include an extra video supplementary material and we will make our code publicly available soon to benefit future works in this field.

\section{Additional Experimental Results}
\label{sec:aer}

\paragraph{Quantitative results for each subject} As aforementioned in the main paper, we compared our approach and state-of-the-art methods for neural human rendering on 6 subjects in the ZJU-MoCap dataset~\cite{peng2021neural}, including \textit{313, 377, 387, 392, 393}, and \textit{394}. Tab.~\ref{tab:novel_view_all} and Tab.~\ref{tab:novel_pose_all} show the detailed quantitative results of the novel view and novel pose scenarios for all subjects. As we can see from the quantitative results, our approach largely outperforms the UV Volumes~\cite{chen2023uv} on all the evaluation metrics. 
Meanwhile, compared to HumanNeRF~\cite{weng2022humannerf}, our approach has shown competitive performance for novel view scenarios, while performing better on novel pose synthesis for all subjects. The quantitative comparisons demonstrate the effectiveness of our approach for neural human reconstruction by learning the canonical neural texture and disentangling the shape and appearance information.

\begin{figure*}[!hbt]
\vspace{-7mm}
\includegraphics[width=\textwidth]{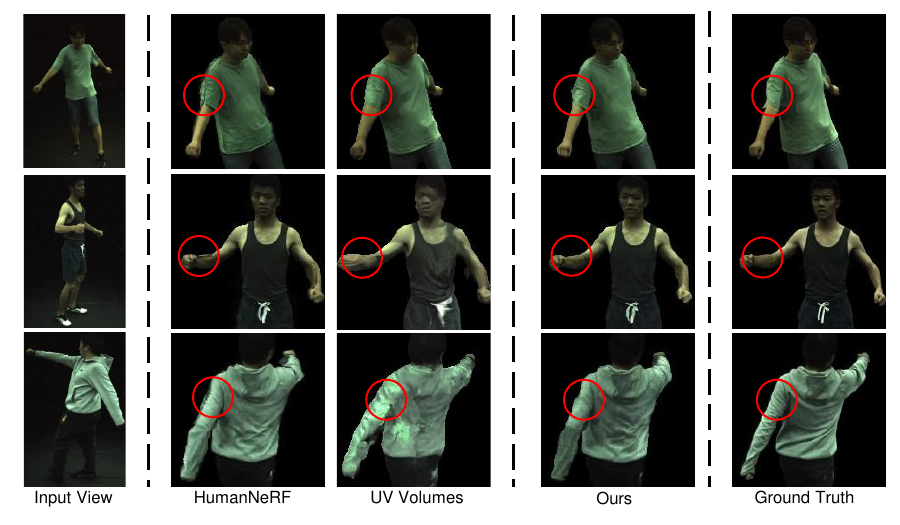}
\vspace{-5mm}
\caption{\textbf{Qualitative results for human rendering on ZJU-MoCap dataset~\cite{peng2021neural} for novel view synthesis.} Our approach gives the best quality for the rendered human body.} 
\label{fig:zju_2}
\vspace{-5mm}
\end{figure*}

\paragraph{Qualitative results for novel view} In Fig.~\ref{fig:zju} 
of the main paper, we have shown the qualitative results for three subjects. We further show the qualitative results of all the rest subjects in Fig.~\ref{fig:zju_2}. HumanNeRF~\cite{weng2022humannerf} generates disconnected body texture with dark lines (see marked regions), and the UV Volumes~\cite{chen2023uv} method suffers from low rendering quality. In contrast, our approach successfully addresses both issues and provides results visually closer to the ground truth.

\begin{table*}[!ht]
\centering
\resizebox{1\textwidth}{!}{
\begin{tabular}{lccccccccr}
\toprule 
\multirow{1}{*}{} & \multicolumn{3}{c}{Subject \textbf{313}} & \multicolumn{3}{c}{Subject \textbf{377}} & \multicolumn{3}{c}{Subject \textbf{387}}\tabularnewline
\cmidrule(lr){2-4}\cmidrule(lr){5-7}\cmidrule(lr){8-10} 
 & PSNR $\uparrow$  & SSIM $\uparrow$  & LPIPS $\downarrow$  & PSNR $\uparrow$  & SSIM $\uparrow$  & LPIPS $\downarrow$  & PSNR $\uparrow$  & SSIM $\uparrow$  & LPIPS $\downarrow$ \tabularnewline
\midrule
UV Volumes~\cite{chen2023uv}  & 24.59  & 0.965  & 0.043  & 23.53  & 0.969  & 0.041  & 21.75  & 0.957  & 0.049 \tabularnewline

HumanNeRF~\cite{weng2022humannerf}  & \textbf{29.57}  & 0.967  & 0.030  & 30.35  & 0.975  & 0.024  & \textbf{27.94}  & \textbf{0.962}  & \textbf{0.037} \tabularnewline

Ours  & 29.40  & \textbf{0.968}  & \textbf{0.029}  & \textbf{30.40} & \textbf{0.977}  & \textbf{0.021}  & 27.73  & \textbf{0.962}  & \textbf{0.037} \tabularnewline
\toprule 
\multirow{1}{*}{} & \multicolumn{3}{c}{Subject \textbf{392}} & \multicolumn{3}{c}{Subject \textbf{393}} & \multicolumn{3}{c}{Subject \textbf{394}}\tabularnewline
\cmidrule(lr){2-4}\cmidrule(lr){5-7}\cmidrule(lr){8-10} 
 & PSNR $\uparrow$  & SSIM $\uparrow$  & LPIPS $\downarrow$  & PSNR $\uparrow$  & SSIM $\uparrow$  & LPIPS $\downarrow$  & PSNR $\uparrow$  & SSIM $\uparrow$  & LPIPS $\downarrow$ \tabularnewline
\midrule 
UV Volumes~\cite{chen2023uv}  & 24.22  & 0.965  & 0.047  & 22.54  & 0.956  & 0.053  & 23.24  & 0.958  & 0.047 \tabularnewline

HumanNeRF~\cite{weng2022humannerf}  & 30.51  & 0.969  & \textbf{0.033}  & \textbf{28.53}  & \textbf{0.961}  & \textbf{0.037}  & \textbf{29.85}  & 0.961  & \textbf{0.036} \tabularnewline

Ours  & \textbf{30.54}  & \textbf{0.970}  & \textbf{0.033}  & 28.34  & \textbf{0.961}  & 0.038  & 29.69  & \textbf{0.962}  & \textbf{0.036}\tabularnewline
\bottomrule
\end{tabular}
}
\vspace{3px}
\caption{\textbf{Novel view synthesis quantitative comparison on ZJU-MoCap dataset.} We show the results of each subject.}
\label{tab:novel_view_all}
\end{table*}

\begin{table*}[!ht]
\resizebox{\textwidth}{!}{
\centering
\begin{tabular}{lccccccccr}
\toprule 
\multirow{1}{*}{} & \multicolumn{3}{c}{Subject \textbf{313}} & \multicolumn{3}{c}{Subject \textbf{377}} & \multicolumn{3}{c}{Subject \textbf{387}}\tabularnewline
\cmidrule(lr){2-4}\cmidrule(lr){5-7}\cmidrule(lr){8-10} 
 & PSNR $\uparrow$ & SSIM $\uparrow$ & LPIPS $\downarrow$ & PSNR $\uparrow$ & SSIM $\uparrow$ & LPIPS $\downarrow $ & PSNR $\uparrow$ & SSIM $\uparrow$ & LPIPS $\downarrow$ \tabularnewline
\midrule 
UV Volumes~\cite{chen2023uv} & 23.36 & 0.968 & 0.043
& 23.56 & 0.979 & 0.035
& 21.97 & 0.962 & 0.045 \\
HumanNeRF~\cite{weng2022humannerf} & 29.85 & 0.970 & 0.027
& \textbf{31.02} & 0.980 & 0.019 
& \textbf{27.87} & \textbf{0.963}& \textbf{0.035} \\
Ours & \textbf{30.36} &\textbf{0.972} & \textbf{0.026}
& 30.98 & \textbf{0.981} & \textbf{0.016}
& 27.49 & \textbf{0.963} & \textbf{0.035} \tabularnewline
\toprule 
\multirow{1}{*}{} & \multicolumn{3}{c}{Subject \textbf{392}} & \multicolumn{3}{c}{Subject \textbf{393}} & \multicolumn{3}{c}{Subject \textbf{394}}\tabularnewline
\cmidrule(lr){2-4}\cmidrule(lr){5-7}\cmidrule(lr){8-10} 
 & PSNR $\uparrow$ & SSIM $\uparrow$ & LPIPS $\downarrow$ & PSNR $\uparrow$ & SSIM $\uparrow$ & LPIPS $\downarrow $ & PSNR $\uparrow$ & SSIM $\uparrow$ & LPIPS $\downarrow$ \tabularnewline
\midrule
UV Volumes~\cite{chen2023uv}  & 25.27 & 0.974 & 0.035
& 22.20 & 0.961 & 0.046 
& 22.02 & 0.961 & 0.043 \\
HumanNeRF~\cite{weng2022humannerf}  & \textbf{31.77} & \textbf{0.975} & \textbf{0.026}
& \textbf{28.84} & \textbf{0.965} & 0.033 
& 28.37 & 0.962 & 0.032 \\
Ours & 31.55 & \textbf{0.975} & \textbf{0.026}
& 28.75 & \textbf{0.965} & \textbf{0.032}
& \textbf{28.82} & \textbf{0.963} & \textbf{0.031} \\
\hline
\end{tabular}
}
\vspace{5px}
\caption{\textbf{Novel pose synthesis quantitative comparison on ZJU-MoCap dataset.} We show the results of each subject.}
\label{tab:novel_pose_all}
\vspace{-14px}
\end{table*}

\begin{figure*}[!hbt]
\includegraphics[width=\textwidth]{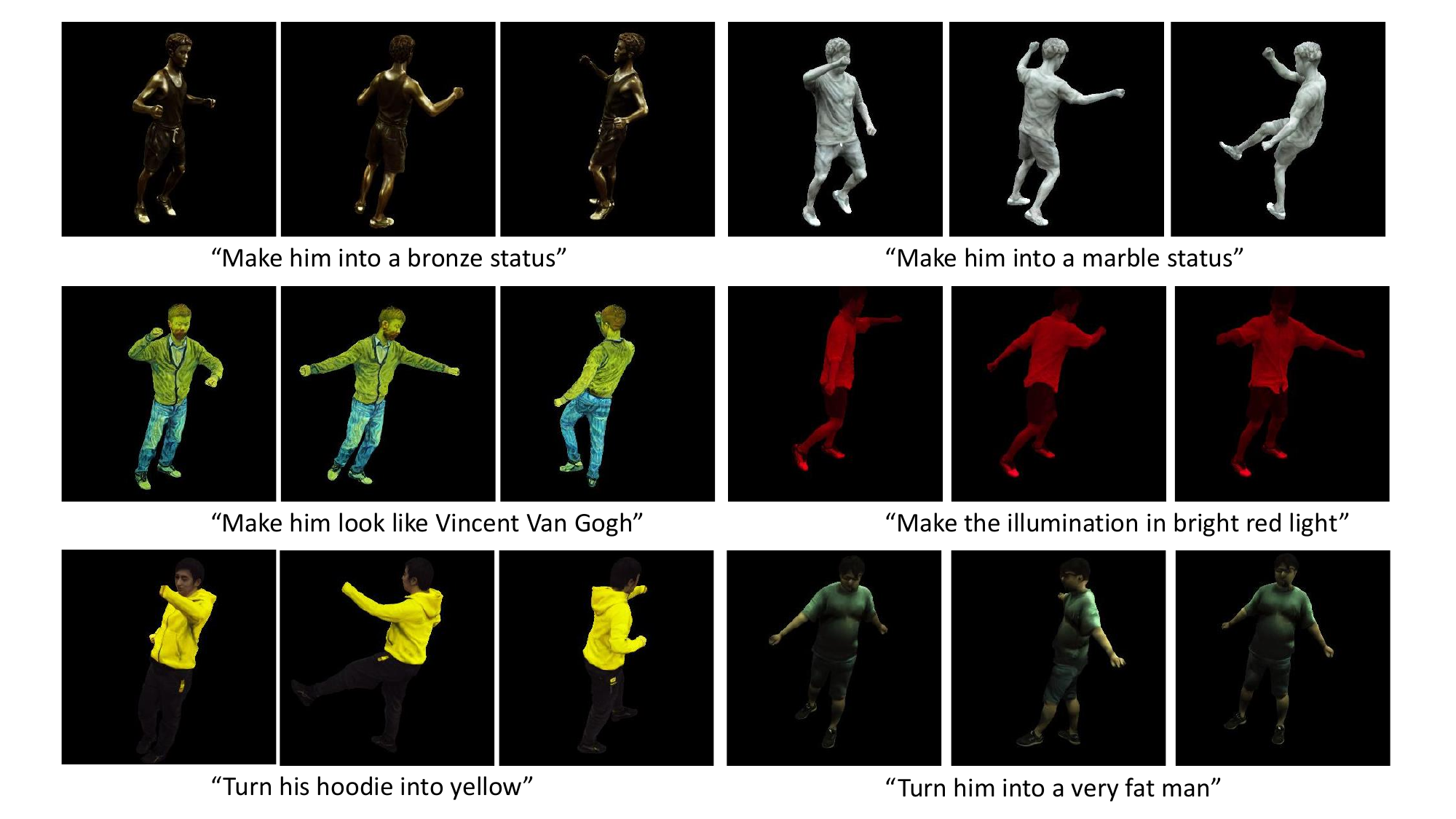}
\vspace{-5mm}
\caption{\textbf{Qualitative results for text Editing for novel view synthesis.} } 
\label{fig:in2n_text}
\end{figure*}

\paragraph{Text Editing Qualitative Assessment for Diverse Subjects.} We present our results on additional subjects in Figure~\ref{fig:in2n_text}, which suggests the versatility of our approach, and showcases its applicability across different subjects.

\section{Additional Implementation Details}
\label{sec:aid}
As described in the main manuscript, our approach reconstructs the human avatar with the disentangled neural feature and enables different manipulations including novel view rendering, novel pose synthesis, and human avatar editing with language brush. 
During the training stage, we first employ the photo-metric loss $\mathcal{L}_{\text{rec}}$ to obtain the densities and colors along the bent rays using the deformation module to establish accurate deformation learning in the initial stage, which is similar to HumanNeRF. We then learn the UVS map and neural texture with the alternative training strategy to optimize regularization loss $\mathcal{L}_{\text{reg}}$ from Eq.~\ref{eq:l_reg} in the main paper and the MSE loss
while freezing the deformation module. Then we use $\mathcal{L}_{\text{rec}}$ with a learning rate $1\times 10^{-4}$ to optimize the rendering details by training UVS map and neural texture jointly. In the end, we further introduce the mask loss and jointly train all the modules for enhancing the neural rendering quality and geometric accuracy. 

When editing with a language brush, the text instruction will guide our model to update the corresponding modules to efficiently edit the shape, illumination, and appearance, where we iteratively supervise our model with training objectives described in Eq.~\ref{eq:l_rec} in the main paper 
from edited ground-truth to fulfill the instructed editing while maintaining the multi-view geometrical consistency. 

\end{document}